\documentclass[runningheads]{llncs}
\usepackage{graphicx}

\usepackage{tikz}
\usepackage{comment}
\usepackage{amsmath,amssymb} 
\usepackage{color}

\usepackage{dsfont}

\newcommand{\etal}{\textit{et al.\ }}
\newcommand{\R}{\mathds{R}}

\usepackage{multirow}
\usepackage{xcolor}

\begin{document}
\pagestyle{headings}
\mainmatter
\def\ECCVSubNumber{5}  

\title{Data augmentation techniques for\\the Video Question Answering task} 

%
\author{Alex Falcon\inst{1,2}\orcidID{0000-0002-6325-9066} \and
Oswald Lanz\inst{1}\orcidID{0000-0003-4793-4276} \and
Giuseppe Serra\inst{2}\orcidID{0000-0002-4269-4501}}
\authorrunning{A. Falcon et al.}
%
\institute{Fondazione Bruno Kessler, Trento 38123, Italy
\\\email{lanz@fbk.eu}\\ \and
University of Udine, Udine 33100, Italy\\
\email{falcon.alex@spes.uniud.it}\\\email{giuseppe.serra@uniud.it}}
\maketitle

\begin{abstract}
Video Question Answering (VideoQA) is a task that requires a model to analyze and understand both the visual content given by the input video and the textual part given by the question, and the interaction between them in order to produce a meaningful answer.
In our work we focus on the Egocentric VideoQA task, which exploits first-person videos, because of the importance of such task which can have impact on many different fields, such as those pertaining the social assistance and the industrial training.
Recently, an Egocentric VideoQA dataset, called EgoVQA, has been released. Given its small size, models tend to overfit quickly. To alleviate this problem, we propose several augmentation techniques which give us a +5.5\% improvement on the final accuracy over the considered baseline. 
\end{abstract}

\section{Introduction}
Video Question Answering (VideoQA) is a task that aims at building models capable of providing a meaningful and coherent answer to a visual contents-related question, exploiting both spatial and temporal information given by the video data. VideoQA is receiving attention from both the Computer Vision and the Natural Language Processing communities, due to the availability of both textual and visual data which require to be jointly attended to in order to give the correct answer \cite{xu2017video,gao2018motion,fan2019heterogeneous}.

Recent advancements in the VideoQA task have also been achieved thanks to the creation of several public datasets, such as TGIF-QA \cite{jang2017tgif} and MSVD-QA \cite{xu2017video}, which focus on web scraped video that are often recorded from a third-person perspective. Even more recently, Fan released in \cite{fan2019egovqa} EgoVQA, an Egocentric VideoQA dataset which provided the basis to study the importance of such task. In fact, several fields can benefit from advancements in the Egocentric VideoQA task: for example, the industrial training of workers, who may require help in understanding how to perform a certain task given what they see from their own perspective; and the preventive medicine field, where Egocentric VideoQA makes it possible to identify sedentary and nutrition-related behaviours, and help elderly people prevent cognitive and functional decline by letting them review lifelogs \cite{doherty2013wearable}. Differently from third-person VideoQA, in the egocentric setting some types of questions can not be {\color{black}posed}, such as those pertaining the camera wearer (\textit{e.g.} ``what am I wearing?''). Moreover, if the question asks to identify an item the camera wearer is playing with, hands occlusion may partially hide the item, making it hard to recognize.

{\color{black}Data augmentation techniques have proven particularly helpful in several Computer Vision tasks, such as image classification \cite{krizhevsky2012imagenet}. Not only they can be helpful to avoid overfitting and thus make the model more general, they can also be used to solve class imbalance in classification problems by synthesizing new samples in the smaller classes \cite{shorten2019survey}.} With respect to third-person VideoQA datasets, EgoVQA is a small dataset comprising around 600 question-answer pairs {\color{black}and the same number of clips}. {\color{black}Since data augmentation is helpful in such contexts but to the best of our knowledge its effectiveness has never been systematically investigated for the Egocentric VideoQA nor for the VideoQA task}, in this work we propose several data augmentation techniques which exploit characteristics given by the task itself. In particular, by exploiting the EgoVQA dataset \cite{fan2019egovqa} we show their impact on the final performance obtained by the ST-VQA model \cite{jang2017tgif}, {\color{black}which is proven to be effective in the study made by Fan}.


The main contributions of this paper can be summarized as follows:
\begin{itemize}
    \item we propose several data augmentation techniques which are purposefully designed for the VideoQA task;
    \item we show the usefulness of our proposed augmentation techniques on the recently released EgoVQA dataset and try to explain why we observe such improvements; 
    \item we achieve a new state-of-the-art accuracy on the EgoVQA dataset;
    \item we will release code and pretrained models to support research in this important field.
\end{itemize}

The rest of the paper is organized as follows: in Section 2, we introduce the related work to the topics involved in this study, namely Egocentric VideoQA and data augmentation techniques; in Section 3, we detail both our proposed augmentation techniques and the architecture we use; Section 4 covers the experiments performed and the discussion of the results that we obtained; finally, Section 5 draws the conclusions of this study.

\section{Related work}
In this section we will discuss the work related to the two main topics involved in our study, \textit{i.e.} Video Question Answering, and data augmentation techniques.

\subsection{VideoQA}
Recently, VideoQA has received a lot of attention \cite{fan2019egovqa,fan2019heterogeneous,gao2018motion,jang2017tgif,xu2017video} from researchers both in Computer Vision and NLP fields. Several reasons can be related to this interest, such as the challenges offered by this task and the availability of several datasets, \textit{e.g.} TGIF-QA \cite{jang2017tgif}, MSRVTT-QA \cite{xu2017video}, MSVD-QA \cite{xu2017video}, ActivityNet-QA \cite{yu2019activitynet}, and TVQA+ \cite{lei2019tvqa+}, populated by many thousands of examples to learn from. 

Modern approaches to this task involve {\color{black}a wide selection of different techniques. Jang \etal proposed in \cite{jang2017tgif} to use both temporal attention and spatial attention, in order to learn which frames and which regions in each frame are more important to solve the task. Later on, attention mechanisms have been also used as a cross-modality fusion mechanism \cite{huang2020location}, and to learn QA-aware representations of both the visual and the textual data \cite{lei2019tvqa+,kim2020dense}. Because of the heterogeneous nature of the appearance and motion feature which are usually extracted from the video clips, Fan \etal \cite{fan2019heterogeneous} also propose to use memory modules, coupled with attention mechanisms, to compute a joint representation of these two types of features. Moreover, to compute the final answer for the given video and question there are multiple approaches. Simpler ones propose to use fully connected networks coupled with non-linear functions \cite{jang2017tgif}, but also more complex solutions have been proposed, \textit{e.g.} based on reasoning techniques which exploit multiple steps LSTM-based neural networks \cite{fan2019heterogeneous} or graphs to better encode the relationships between the visual and textual data \cite{huang2020location,jiang2020reasoning}.

Finally, given the multitude of VideoQA datasets, there can also be multiple types of information to exploit. In fact, not only clips, questions, and answers are exploited to solve this task: as an example, TVQA+ \cite{lei2019tvqa+} also provides subtitles and bounding boxes, by using which it is possible to improve the grounding capabilities of the VideoQA model in both the temporal and the spatial domain.}

\subsection{Egocentric VideoQA}

{\color{black}On the other hand, Egocentric VideoQA was a completely unexplored} field until very recently, {\color{black}when Fan released the EgovQA dataset in \cite{fan2019egovqa}. Yet, considering the recent advancements in several fields of the egocentric vision, such as action recognition and action anticipation \cite{Damen2020RESCALING,Damen2020Collection}, Egocentric VideoQA also plays a primary role in the understanding of the complex interactions of the first-person videos.}

Both VideoQA and Egocentric VideoQA usually deal with two main types of tasks: the ``open-ended'' and the ``multiple choice'' task \cite{fan2019egovqa,jang2017tgif,xu2017video,yu2019activitynet}. Given a visual contents-related question, the difference between the two is due to how the answer is chosen: in the former, an answer set is generated from the most frequent words (\textit{e.g.} top-1000 \cite{xu2017video,yu2019activitynet}) in the training set and the model needs to choose the correct answer from it, \textit{i.e.} it is usually treated as a multi-class classification problem; in the latter the model needs to select the correct answer from a small pool of candidate answers (\textit{e.g.} five choices \cite{fan2019egovqa,jang2017tgif}), which are usually different for every question. In this work we focus on the multiple choice task.

{\color{black}Together with the release of the EgoVQA dataset, Fan also provided} in \cite{fan2019egovqa} a baseline made of four models borrowed from the VideoQA literature \cite{fan2019heterogeneous,gao2018motion,jang2017tgif}. These models use the same backbone, which consists in a frozen, pretrained VGG-16 \cite{simonyan2014very} to extract the frame-level features; a frozen, pretrained C3D \cite{tran2015learning} to extract the video-level features; and a pretrained GloVe \cite{jeffreypennington2014glove} to compute the word embeddings. The four models can be seen as extensions of a basic encoder-decoder architecture (referred to as ``ST-VQA without attention'' in \cite{fan2019egovqa}): ``ST-VQA with attention'' is based on \cite{jang2017tgif} and uses a temporal attention module to attend to the most important frames in the input clip; ``CoMem'' is based on \cite{gao2018motion} and involves the usage of two memory layers to generate attention cues starting from both the motion and appearance features; and finally ``HME-VQA'' \cite{fan2019heterogeneous} uses two heterogeneous memory layers and a multi-step reasoning module. {\color{black}In \cite{fan2019egovqa}, Fan shows that these four models achieve similar performance despite the introduction of several cutting-edge modules. Because of this reason and because of its simplicity, in this work we focus on the ``ST-VQA with attention'' model.}

\subsection{Data augmentation techniques}
{\color{black}Several Computer Vision tasks, such as image classification \cite{krizhevsky2012imagenet} and handwritten digit classification \cite{lecun1998gradient}, have seen great improvements by exploiting data augmentation techniques, through which the size of the training set can be expanded artificially by several orders of magnitude. This leads to models which are far less susceptible to overfitting and more prone to give better results during the testing phase.}

{\color{black}Whereas papers about first- or third-person VideoQA never mention any augmentation technique,} in the VisualQA task, which deals with question-answering over images, there are some papers which try to tackle this opportunity by exploiting template-based models or generative approaches. Using a semantic tuple extraction pipeline, Mahendru \etal \cite{mahendru2017promise} extract from each question a premise, \textit{i.e.} a tuple made of either an object, or an object and an attribute, or two objects and a relation between them, from which new question-answer pairs are constructed using previously built templates. Kafle \etal \cite{kafle2017data} proposes two techniques. The first is a template-based method which exploits the COCO dataset \cite{lin2014microsoft} and its segmentation annotations to generate new question-answer pairs of four different types, exploiting several different templates for each type. The second approach is based on sequentially generating new questions (and related answers) by conditioning an LSTM-based network on the image features from the ``VQA dataset'' \cite{antol2015vqa}. 
Both these methods focus on creating new question-answer pairs for the same image, either by exploiting purely linguistic aspects or by using visual information to better guide the generation. {\color{black}Yet, as the authors report in the respective papers, these methods although reliable are not error-free \cite{mahendru2017promise,kafle2017data}; on the other hand, our proposed techniques are simple yet effective and they do not raise issues. In particular, in our study we focus on exploiting both the visual and the textual data,} although we do not create new questions: two of our techniques create new candidate answers for the same question to strengthen the understanding of the concepts contained in the question and to better distinguish between the correct and the wrong answers, {\color{black}and the other technique creates ``new'' clips by horizontally flipping the frames and consistently updating both the question and the candidate answers accordingly}. 

\section{Methodology}
\begin{figure*}[!t]
\centering
\includegraphics[width=4.75in]{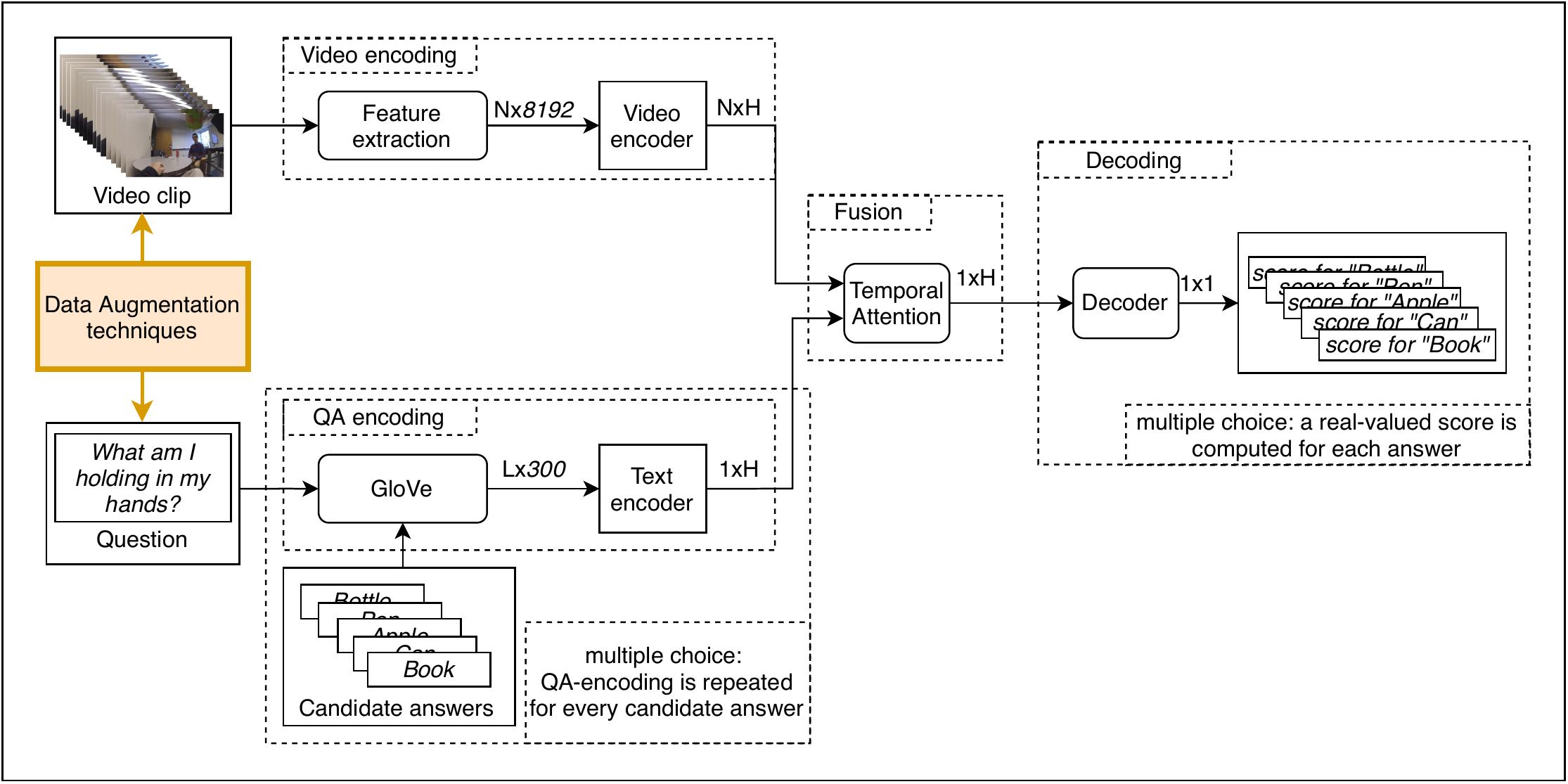}
\caption{Model used in our study. The feature extraction module is made of VGG and C3D, whose output consists of $N$ frames and 8192 features. The output of GloVe consists in fixed-length vectors (\textit{i.e.} embeddings) of embedding size $E=300$. The Video Encoder and the Text Encoder have a similar structure, but whereas the output of the former is a sequence (length $N$) of hidden states, the output of the latter is a single hidden state. The Decoder outputs a single real-valued score for each candidate answer.}
\label{fig:arch}
\end{figure*}

In this section we will introduce and describe the proposed augmentation techniques. Moreover, we will also discuss {\color{black}and describe the model used in our study, which is called ``ST-VQA'' and was initially introduced by Jang \etal in \cite{jang2017tgif}.} 

\subsection{Augmentation techniques}
{\color{black}Following the work made in \cite{fan2019egovqa}, we are working on the multiple choice setting. For each video and question five candidate answers are provided, of which only one is correct. 
The wrong answers are randomly sampled from a candidate pool based on the question type, \textit{i.e.} if the question requires to recognize an action, the five candidate answers (both the right one and the four wrong) are actions. By doing so, the model is encouraged to understand the visual contents in order to reply to the question, avoiding the exploitation of pure textual information (\textit{e.g.} exploiting the question type to filter out some of the candidate answers).}

We propose to use {\color{black}three} {\color{black}simple} augmentation techniques designed for the VideoQA task {\color{black} and which exploit the multiple choice setting}: resampling, mirroring, {\color{black}and horizontal flip.} This is not only helpful when dealing with the overfitting, but can also give the model a better understanding of what the questions is asking for and {\color{black}make the model more robust to variations in the input frames}. 

\subsubsection{Resampling}
\begin{figure}[h]
    \centering
    \includegraphics[width=330pt]{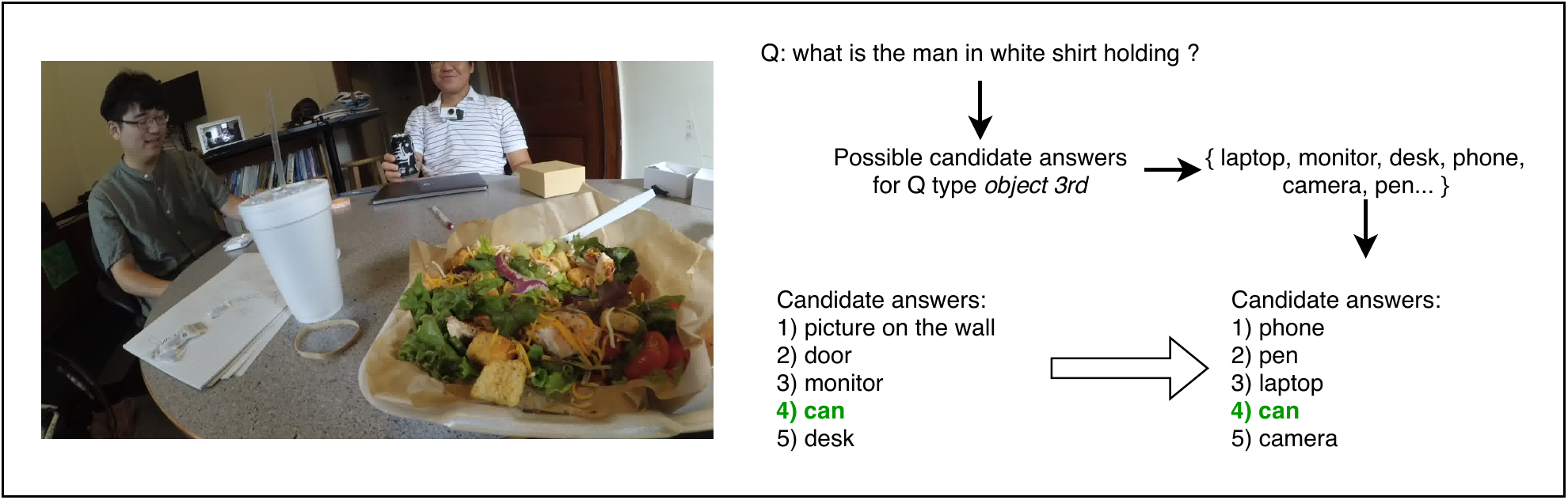}
    \caption{Example of the ``resampling'' technique applied to a video clip in the EgoVQA dataset.}
    \label{fig:eg_resamp}
\end{figure}

Given a question Q, in the multiple choice setting a handful (\textit{e.g.} 5 choices \cite{fan2019egovqa}, \cite{jang2017tgif}) of candidate answers are considered. The first technique consists in fixing the correct answer and then \textit{resampling} the wrong ones. By doing so, using the same video and question, we can show the model several more examples of what is \textit{not} the correct answer. This should give the model the ability to better distinguish what the question is and is not asking for. An example is shown in Figure \ref{fig:eg_resamp}.

Considering that the amount of possible tuples of wrong answers is exponentially big, we are restricting the pool of wrong answers to those pertaining the same question type of Q. Moreover, we are not considering all the possible tuples in the pool.

\subsubsection{Mirroring}
\begin{figure}[h]
    \centering
    \includegraphics[width=330pt]{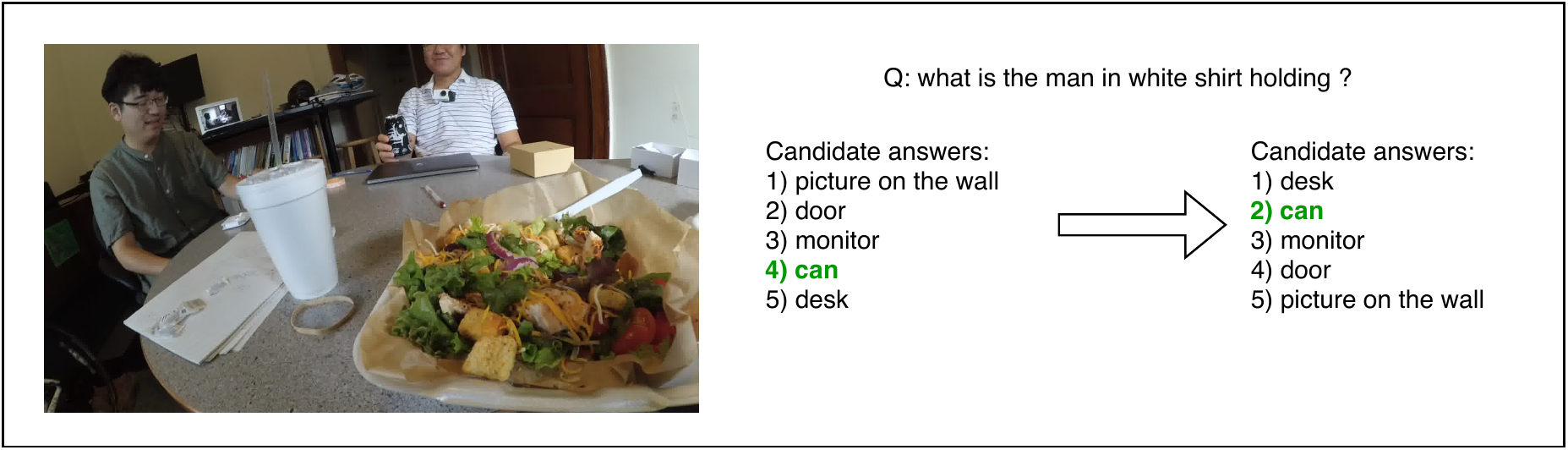}
    \caption{Example of the ``mirroring'' technique applied to a video clip in the EgoVQA dataset.}
    \label{fig:eg_mirror}
\end{figure}

Given a question, it may be that the correct answer in the rows of the dataset is often placed in the same position. This can create biases in the model which may tend to prefer an answer simply based on its position (w.r.t. the order of the candidates). To relieve some of this bias we propose the \textit{mirroring} technique, which consists in simply adding a row to the dataset where the order of the candidate answers (and the label value) is mirrored. An example is shown in Figure \ref{fig:eg_mirror}.

\subsubsection{{\color{black}Horizontal flip}}
\begin{figure}[h]
    \centering
    \includegraphics[width=330pt]{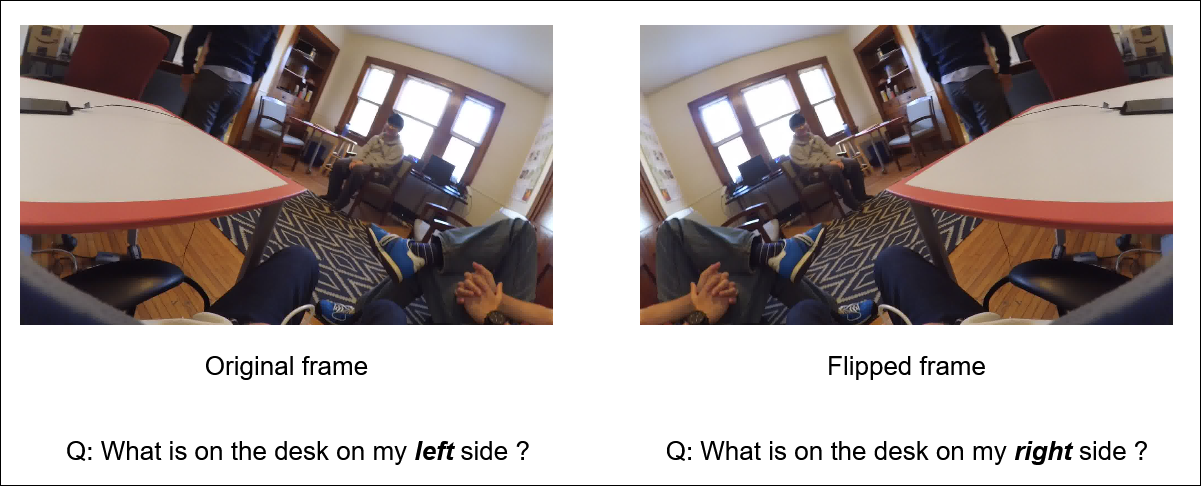}
    \caption{Example of the ``horizontal flip'' technique applied to a frame of a video clip in the EgoVQA dataset.}
    \label{fig:hflip}
\end{figure}

One of the most common image data augmentation techniques consists in horizontally flipping the images, which often improves the model performance thanks to the availability of newly created images which are taken both from the left and from the right. This technique may prove useful in a VideoQA setting as well, but it should not be applied lightly because it is a non-label preserving transformation: horizontally flipping the considered frame means that an object which was on the left side of the frame appears on the right side after the transformation, and viceversa, eventually creating wrong labels if not updated correctly. Thus, when flipping the frames in the video clip both the question and the candidate answers likely need to be updated (\textit{e.g.} Figure \ref{fig:hflip}).

\subsection{QA encoding: Word embedding and Text Encoder} 
{\color{black}As shown in Fig. \ref{fig:arch}, it can be seen as made of four blocks: Question-Answer (QA) Encoding, Video Encoding, Fusion, and Decoding. }
To compute the word embeddings for the question and the answers, we consider GloVe \cite{jeffreypennington2014glove}, pretrained on the Common Crawl dataset\footnote{The Common Crawl dataset is available at http://commoncrawl.org}, which outputs a vector of size $E=300$ for each word in both the question and the answers. Since GloVe is not contextual, question and answer can be given in input to the model either separately or jointly obtaining the same embedding.

First of all, the question and the candidate answer are tokenized, \textit{i.e.} they are split in sub-word tokens and then each of them receives an identifier, based on the vocabulary used by GloVe. Let $q_1 \dots q_m$ and $a_1 \dots a_n$ be the sequence of $m$ words of the question and $n$ words of (one of the candidate) answer, and let $L=m+n$. Thus, let $\phi_{q} \in \R^{m \times E}$ be the question embedding, and $\phi_{a} \in \R^{n \times E}$ be the answer embedding. The final question-answer embedding is computed as their concatenation, \textit{i.e.} $\phi_{w} = [\phi_{q}, \phi_{a}] \in \R^{L \times E}$.

Then the Text Encoder, consisting of two stacked LSTM networks, is applied to $\phi_{w}$. By concatenating the \textit{last} hidden state of both the LSTM networks we obtain the encoded textual features $\epsilon_{w} \in \R^{1 \times H}$, where H is the hidden size.

\subsection{Video Encoding}
From each input video clip, both motion and appearance features are obtained in the Video Encoding module. In particular, the appearance features are computed as the \textit{fc7} activations ($\phi_{a} \in \R^{N \times 4,096}$) extracted from a frozen VGG-16 \cite{simonyan2014very}, pretrained on ImageNet \cite{krizhevsky2012imagenet}. {\color{black}We use VGG because we want to keep the positional information extracted by the convolutional layers, which would be otherwise lost in deeper networks, such as ResNet \cite{he2016deep}, that exploit a global pooling layer before the FC layers.} Similarly, the motion features are computed as the \textit{fc7} activations ($\phi_{m} \in \R^{N \times 4,096}$), extracted from a frozen C3D \cite{tran2015learning}, pretrained on Sports1M \cite{karpathy2014large} and fine-tuned on UCF101 \cite{soomro2012ucf101}. Finally we concatenate these features and obtain a feature vector $\phi_{a, m} \in \R^{N\times8,192}$, which is then encoded by a Video Encoder module, consisting of two stacked LSTM networks. The only difference between the Text and Video Encoder module is that the output $\epsilon_{v}$ of the latter consists in the concatenation of the full sequence of hidden states from both the networks, and not only the last hidden state. Thus $\epsilon_{v} \in \R^{N \times H}$ represents the encoded video features. 

\subsection{Fusion}
Depending on the question (and eventually the candidate answer), a frame may be more or less relevant. To try and exploit this information, the fusion block consists of a temporal attention module that lets the model learn automatically which frames are more important based on both the encoded video features and the textual features. In particular, the temporal attention module is based on the works by Bahdanau \etal \cite{bahdanau2014neural} and by Hori \etal \cite{hori2017attention}. It receives in input the encoded video features $\epsilon_{v}$ and the encoded textual features $\epsilon_{w}$, and can be described by the following equations:
\begin{equation}
\label{eq:tpatt3}
    \omega_{s} = tanh(\epsilon_{v} W_v + \epsilon_{w} W_w + b_s) W_s
\end{equation}
\begin{equation}
\label{eq:tpatt4}
    \alpha = softmax(\omega_{s})
\end{equation}
\begin{equation}
\label{eq:tpatt5}
    \omega_{a} = \mathds{1} (\alpha \circ \epsilon_{v})
\end{equation}
where $W_v, W_w \in \R^{H \times h}$, $W_s \in \R^{h}$ are learnable weight matrices, $b_s \in \R^{1 \times h}$ is a learnable bias. $\mathds{1}$ is a row of ones ($1^{1 \times N}$). $\circ$ represents the element-wise multiplication operator. The output of the Fusion module is a feature vector $\omega_a \in \R^{1 \times H}$.

\subsection{Decoding}
Finally, the decoding step considers both the attended features computed by the Fusion block and the encoded textual features, as proposed by Fan \cite{fan2019egovqa}. In our multiple choice setting, the decoding is performed five times, \textit{i.e.} for each Q-A pair, with different textual features producing five different scores, one per candidate answer. It can be described by the following equations:
\begin{equation}
\label{eq:9}
    d_f = tanh(\omega_{a} W_a + b_a)
\end{equation}
\begin{equation}
\label{eq:10}
    d_r = (d_f \circ \epsilon_{w}) W_d + b_d
\end{equation}
where $W_a \in \R^{H \times H}$ and $W_d \in \R^{H \times 1}$ are learnable weight matrices, $b_a \in \R^{1 \times H}$ and $b_d \in \R$ are learnable biases, $d_f \in \R^{1 \times H}$, $d_r \in \R$. $d_r$ can be seen as the score obtained by testing a specific candidate answer (out of the five possible choices related to the given question).

\subsection{Loss function}
The model is trained using a pairwise hinge loss, as is done in \cite{fan2019egovqa,jang2017tgif}. The loss function can be defined as follows:
\begin{equation}
\label{eq:loss1}
\mathcal{L}_{c, r} = 
    \begin{cases}
    0 & \text{if } c = r \\
    max(0, 1 + s_n - s_p) &  \text{if } c \ne r 
    \end{cases}
\end{equation}
where $s_n$ and $s_p$ are respectively the scores $d_f$ computed by the decoder for the choice $c \in \{1, \dots, 5\}$ and the right answer $r$.
\begin{equation}
\label{eq:loss2}
\mathcal{L} = \sum_{q \in \mathcal{Q}} \sum_{c=1}^{5} \mathcal{L}_{c, r}
\end{equation}
Here $\mathcal{Q}$ is the set of the questions, and $r$ is the right answer for the question $q$.

\begin{figure*}[!t]
\centering
\includegraphics[width=330pt]{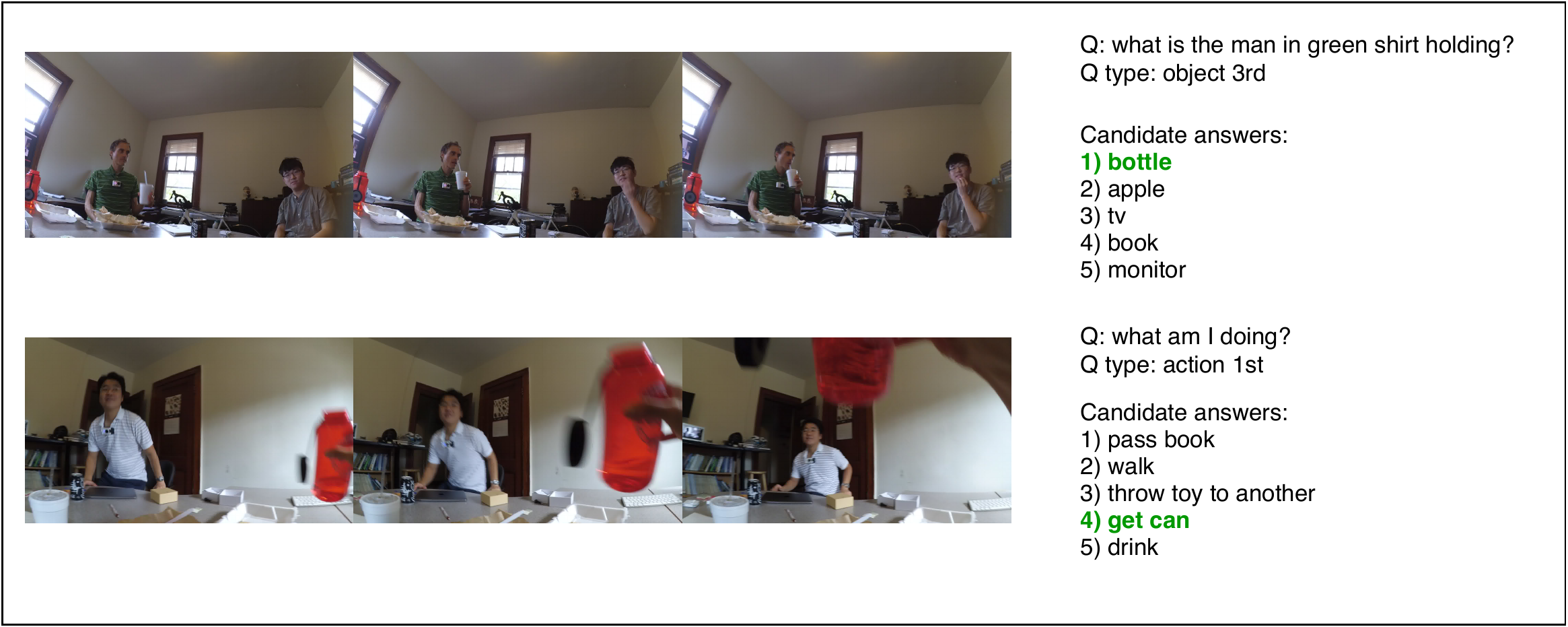}
\caption{Samples of video clips, questions, and candidate answers from the EgoVQA dataset.}
\label{fig:egovqa_sample}
\end{figure*}

\section{Results}
In this section we {\color{black}briefly describe} the dataset used to perform the experiments, and we discuss both the overall results and the per question type results.

\subsection{EgoVQA dataset}
The EgoVQA dataset was recently presented by Fan \cite{fan2019egovqa}. It features more than 600 QA pairs and the same number of clips, which are 20-100 seconds long and are obtained by 16 egocentric videos (5-10 minutes long) based on 8 different scenarios. An example of these egocentric videos and QA pairs can be seen in Fig. \ref{fig:egovqa_sample}. The questions can be grouped in eight major types and they are described in Table \ref{tab:questions}.

\begin{table}[t]
\centering
\caption{Description of the question types available for testing in the EgoVQA dataset.} 
\label{tab:questions}
\begin{tabular}{llll}
\hline
Code & Question type & Quantity & Example \\
\hline
$Act_{1st}$ & Action 1st & 67 & ``what am I doing'' \\
$Act_{3rd}$ & Action 3rd & 108 & ``what is the man in red clothes doing'' \\
$Obj_{1st}$ & Object 1st & 54 & ``what am I holding in my hands'' \\ 
$Obj_{3rd}$ & Object 3rd & 86 & ``what is placed on the desk'' \\ 
$W\!ho_{1st}$ & Who 1st & 13 & ``who am I talking with'' \\  
$W\!ho_{3rd}$ & Who 3rd & 63 & ``who is eating salad'' \\
$Cnt$ & Count & 64 & ``how many people am I talking with'' \\
$Col$ & Color & 31 & ``what is the color of the toy in my hands'' \\
\hline
\end{tabular}
\end{table}


\subsection{Implementation details}
In our setting, we fixed $H = 512$ and $h = 256$. To optimize the parameters we used the Adam \cite{kingma2014adam} optimizer with a fixed learning rate of $10^{-3}$ and a batch size of 8. To implement our solution we used Python 2.7, Numpy 1.16, and PyTorch 1.4. {\color{black}A PyTorch implementation will be made available to further boost the research in this important area at https://github.com/aranciokov/EgoVQA-DataAug}.

\subsection{Discussion of the results}
Table \ref{tab:res} shows the results obtained for each of the three splits proposed in \cite{fan2019egovqa} by applying, with different combinations, our proposed augmentation techniques. Table \ref{tab:qtres} shows the results based on the question type, whose details (and codes, such as ``$Act_{1st}$'' and ``$Act_{3rd}$'') are {\color{black}defined} in Table \ref{tab:questions}. 
\begin{table}[]
    \centering
    \caption{Per-question type results obtained by applying the proposed techniques on the EgoVQA dataset.}
    \begin{tabular}{lccccc}
    \hline
    \multicolumn{1}{c}{\multirow{2}{*}{Augmentation}} &     \multicolumn{3}{c}{Accuracy (\%) on split} & \\ \cline{2-4}
    \multicolumn{1}{c}{} & 0 & 1 & 2 & Avg \\ \hline
    ST-VQA \cite{jang2017tgif}  & 31.82 & 37.57 & 27.27 & 32.22  \\
    \hline
    + mirroring & 32.58 & 40.46 & 23.53 & 32.19 \\
    + resampling & 26.52 & 28.90 & 29.41 & 28.28 \\
    \, \, + mirroring & 37.88 & 36.42 & \textbf{30.48} & 34.93 \\
    {\color{black}+ horizontal-flip} & 34.09 & 41.62 & 25.13 & 33.61 \\
    \, \, + resampling & 37.12 & 35.26 & 25.67 & 32.68 \\
    \, \, \, \, + mirroring & \textbf{40.91} & \textbf{43.35} & 28.88 & \textbf{37.71} \\
    \hline
    \end{tabular}
    \label{tab:res}
\end{table}

Overall it can be seen that, when used in conjunction, the proposed augmentation techniques help improving the performance obtained by the considered model. 

Looking at the results per question type, it is possible to notice that: \begin{itemize}
    \item the ``resampling'' technique is particularly helpful when it comes to counting objects (``$Cnt$'') and identifying objects used by actors in front of the camera wearer (``$Obj_{3rd}$'');
    \item the ``mirroring'' technique shows sensible improvements during the identification of actors, both when the camera wearer is interacting with them (``$W\!ho_{1st}$'') and when they are performing certain actions in front of the camera wearer itself (``$W\!ho_{3rd}$'').
    \item {\color{black}the ``horizontal flip'' technique} is especially helpful when the model needs to identify the actions performed by (``$Act_{1st}$'') and the objects over which the action is performed by the camera wearer (``$Obj_{1st}$''). Moreover, it gives the model a great boost in recognizing colors (``$Col$'').
\end{itemize}

\begin{table}[]
    \centering
    \caption{Per-question type results obtained by applying the proposed techniques on the EgoVQA dataset.}
    \begin{tabular}{lcccccccc}
    \hline
    \multirow{2}{*}{Augmentation} & \multicolumn{7}{c}{Question type accuracy (\%)}        \\ \cline{2-9}
    \multicolumn{1}{c}{} & $Act_{1st}$ & $Act_{3rd}$ & $Obj_{1st}$ & $Obj_{3rd}$ & $W\!ho_{1st}$ & $W\!ho_{3rd}$ & $Cnt$ & $Col$ \\ \hline
    ST-VQA \cite{jang2017tgif} & 28.36 & 30.56 & 31.48 & 31.40 & 46.15 & 34.92 & 35.94 & 32.26   \\ \hline
    + mirroring & 26.87 &  33.33 &  35.19 &  27.91 &  \textbf{53.85} &  46.03 &  26.56 &  19.35 \\
    + resampling & 26.87 &  26.85 &  20.37 &  37.21 &  15.38 &  23.81 &  \textbf{43.75} &  16.13 \\
    \, \, + mirroring & 25.37 &  34.26 &  25.93 &  41.86 &  15.38 &  \textbf{47.62} &  42.19 &  19.35 \\
    {\color{black}+ horizontal-flip} & \textbf{40.30} &  36.11 &  \textbf{42.59} &  25.58 &  38.46 &  28.57 &  26.56 &  \textbf{41.94} \\
    \, \, + resampling & 34.33 &  37.96 &  22.22 &  33.72 &  15.38 &  33.33 &  29.69 &  29.03 \\
    \, \, \, \, + mirroring & 31.34 &  \textbf{39.81} &  22.22 &  \textbf{44.19} &  15.38 &  \textbf{47.62} &  37.50 &  38.71 \\
    \hline
    \end{tabular}
    \label{tab:qtres}
\end{table}
Both in the questions of type ``$Cnt$'' and ``$Obj_{3rd}$'' the model is required to recognize an object: in fact, whereas in the latter the model needs to identify an object by distinguishing among the five candidates, the former also requires the model to understand what such object is in order to count how many times it occurs in the scene. It is interesting to notice that in our ``resampling'' technique we are not augmenting the questions of type ``$Cnt$'', because the only five possible candidate answers in the dataset for such question type are the numbers from ``one'' to ``five''. Thus, since we are able to observe this improvement in both these question types, it likely implies that such data augmentation technique helps the model to better distinguish among the different objects available in the dataset because it provides several more examples where the model needs to understand which object is the right one among several (wrong) candidate answers. 

In the case of the question types ``$W\!ho_{3rd}$'' and ``$Act_{3rd}$'' the improved performance {\color{black}may be} due to two aspects: first of all, since they both require to recognize actions performed by actors in front of the camera wearer, the accuracy gain obtained in one type transfers (to some extent) to the other type, and viceversa; and then to the ``mirroring'' technique, since it is possible to observe that in the training set there is a bias in both question types towards one of the last two labels. In particular, over the three training splits, the last candidate answer is the correct one 60 times over 203 questions (29.55\%) of type ``$Act_{3rd}$'', whereas ``$W\!ho_{3rd}$'' counts 27 instances of the second-to-last candidate answer {\color{black}over a total of 77 questions} (35.06\%). Using the ``mirroring'' technique it is thus possible to reduce this bias, making the model more robust. {\color{black} It is interesting to notice that in both these question types, the addition of the ``horizontal flip'' technique gives a further boost in the accuracy of the model. This is likely related to the fact that several questions in the training data also contain a positional information (``left'', ``right'') of the actor involved: in particular, for the type ``$Act_{3rd}$'' there are respectively 16 and 20 questions mentioning ``left'' or ``right''} over a total of 203 {\color{black}questions, whereas for ``$W\!ho_{3rd}$'' there are respectively 2 and 3 over a total of 77.}

In the question type ``$W\!ho_{1st}$'' we can observe a sensible improvement with the ``mirroring'' technique. Although the reason are likely similar (considering that the first candidate answer is the right one 11 times over 20 instances for ``$W\!ho_{1st}$''), we prefer not to make any {\color{black}conclusive} claim given that there is only a total of 20 instances in the training set and 13 in the testing set.

{\color{black}The ``horizontal flip'' technique shines when asked to recognize which object the camera wearer is interacting with (``$Obj_{1st}$'') and to identify which action (``$Act_{1st}$'') is performed by the camera wearer itself. The improvement} over the former {\color{black}question type may be explained by the fact that several of its questions in the training data involve a positional information: in particular, there are respectively 18 and 9 questions containing ``left'' or ``right'' over a total of 86 questions. On the other hand, the great improvement in the latter question type (whose questions are almost all of the form ``what am I doing'') is likely justifiable by the greater amount of different visual data available for training.}

Finally, {\color{black}among} our proposed techniques{\color{black}, only the ``horizontal flip'' seems} to cope well with ``$Col$'' questions. This question type is particularly tough because it requires the model to recognize the object which the question is referring to, the action which is performed over the object (35/49 total instances), and sometimes even the colors of the clothes of the actor (\textit{e.g.} ``what is the color of the cup held by the man in \textit{black} jacket'', 13/49 total instances). {\color{black}First of all}, the ``mirroring'' technique does not help: the training split are slightly biased towards the first and the last labels (respectively, 10 and 16 over 49 instances), meaning that in this case the proposed technique {\color{black}does not resolve the bias} towards these two labels. 
{\color{black}Secondly}, our ``resampling'' technique is not helping because there are only six unique colors in the dataset, thus it is not creating enough new rows. {\color{black}Thirdly, only 2 over a total of 54 questions of this type in the training data contain ``left'' or ``right'', likely implying that the improvement obtained by the ``horizontal flip'' technique is due to having more visual data which forces the model to better understand where to look for the object targeted by the question.}

\section{Conclusion}
Egocentric VideoQA is a task introduced recently in \cite{fan2019egovqa} which specializes the VideoQA task in an egocentric setting. It is a challenging task where a model needs to understand both the visual and the textual content of the question, and then needs to jointly attend to both of them in order to produce a coherent answer. In this paper we propose several data augmentation techniques purposefully designed for the VideoQA task. The ``mirroring'' technique tries to partially remove the ordering bias in the multiple choice setting. The ``resampling'' technique exploits the training dataset to create new question-answer pairs by substituting the wrong candidate answers with different candidates from the same question type, in order to feed the network with more examples of what is not the target of the question. {\color{black}Finally, the ``horizontal-flip'' technique exploits both the visual and the textual content of each row in the dataset, and aims at giving the model the ability to differentiate between ``left'' and ``right''.} To show the effectiveness of these techniques, we test them on the recently released EgoVQA dataset and show that we are able to achieve a sensible improvement {\color{black}(+5.5\%)} in the accuracy of the model.

{\color{black}As a future work, we are both considering to explore our proposed augmentation techniques with other architectures, such as the HME-VQA model \cite{fan2019heterogeneous}, and to replicate these experiments in third-person VideoQA datasets. Moreover, we are considering several different augmentation techniques that deal with the linguistic aspects and the visual information both separately and jointly. In particular, we think that considering them jointly is of most interest because of the inherent characteristics of the problem setting, which requires the model to understand both linguistic and visual clues together. A purely linguistics technique which we plan to explore consists in a variation of the ``mirroring'' technique which \textit{permutes} the candidate answers instead of simply mirroring them: this should reduce the ordering bias in all the possible situations, even those where the ``mirroring'' technique is weaker. Then, considering that the egocentric camera may not be aligned at all times due to the camera wearer moving in the scenario, a simple technique which might help consists in applying small rotations to the video clips, similarly to what is done in image tasks. Finally, reversing the video clips and updating both question and the candidate answers accordingly (\textit{e.g.} by ``reversing'' the name of the actions performed in the video clip) may give the model a more clear understanding of the actions, while better exploiting the sequential nature of the visual data.}



\clearpage
%
%
\bibliographystyle{splncs04}
\bibliography{main}
\end{document}